\documentclass[letter, 10pt, conference]{ieeeconf}
\IEEEoverridecommandlockouts
\usepackage{cite}

\usepackage{amsmath,amssymb,amsfonts}
\usepackage{algorithmic}
\usepackage{graphicx}
\usepackage{textcomp}
\usepackage{xcolor}
\def\BibTeX{{\rm B\kern-.05em{\sc i\kern-.025em b}\kern-.08em
    T\kern-.1667em\lower.7ex\hbox{E}\kern-.125emX}}
\begin{document}

\title{Learning Behavior Trees with Genetic Programming in Unpredictable Environments\\
% Updated thanks to conform to requirements from SSF: "This project is financially supported by the Swedish Foundation for Strategic Research"
\thanks{This project is financially supported by the Swedish Foundation for Strategic Research and by the Wallenberg AI, Autonomous Systems, and Software Program (WASP) funded by the Knut and Alice Wallenberg Foundation. The authors gratefully acknowledge this support.}
}

\author{\authorblockN{Matteo Iovino\authorrefmark{1}\authorrefmark{3}, Jonathan Styrud\authorrefmark{2}\authorrefmark{3}, Pietro Falco\authorrefmark{1} and Christian Smith \authorrefmark{3}}\\
\thanks{\authorrefmark{1}ABB Corporate Research, Västerås, Sweden}
\thanks{\authorrefmark{2}ABB Robotics, Västerås, Sweden}
\thanks{\authorrefmark{3}Division of Robotics, Perception and Learning, Royal Institute of Technology (KTH), Stockholm, Sweden}
}

\maketitle

\begin{abstract}
Modern industrial applications require robots to be able to operate in unpredictable environments,
%It is also desirable that new robot policies or 
and programs to be created with a minimal effort, as there may be frequent changes to the task. In this paper, we show that genetic programming can be effectively used to learn the structure of a behavior tree (BT) to solve a robotic task in an unpredictable environment. Moreover, we propose to use a simple simulator for the learning and demonstrate that the learned BTs can solve the same task in a realistic simulator, reaching convergence without the need for task specific heuristics. %This is achieved by synthesizing BTs which are naturally reactive to failures and including the success probability in the fitness function, driving towards solutions that avoid actions that are more likely to fail. Finally, we show that
The learned solution is tolerant to faults%on the manipulation, localization and navigation actions
, making our method appealing for real robotic applications. 
\end{abstract}


\section{Introduction}

%\begin{itemize}
%    \item Learning BT: describe problem and complications (- computation time)
%    \item \textbf{Contribution 1:} GP can learn BTs in a probabilistic environment making the solution fault tolerant
%    \item \textbf{Contribution 2:} it is possible to use a state machine simulator to learn faster
%\end{itemize}

Modern industrial applications, with robots sharing environments with humans, require robots to be able to operate in unpredictable environments. This can be achieved by controlling the robot with a policy, rather than a prescribed sequence of actions, to support handling unexpected outcomes of robot actions, or different types of faults and errors. For modern, flexible, manufacturing environments, it is also desirable that new robot policies or programs can be created with a minimum of effort, as changing requirements may require frequent changes to the task the robot should perform.

%\todo{... rather than XXX - Christian}
%\todo{It would be good to elaborate a bit more this point, as planners can be used in unstructured environments anyway- Pietro - I expand this talking about limitations of planners in related work, so I removed it from here}. 
In this paper, we propose a method to learn a policy to solve a mobile manipulation task in unpredictable environments. We use Behavior Trees (BTs) to represent the policy, due to their reactivity and modularity, and Genetic Programming (GP) for the learning, as it is a good fit for the modularity and the structure of a BT.  %\todo{Motivate why?}
The first contribution of this paper is to show that GP can be used to learn a BT which is robust to faults, both reactively and proactively, favoring designs that minimize the risk of error.
%\todo{  Could we emphasize also that it handle faults when they happen (robustness to fault) but also try to avoid faults proactively, as in the example of the safe and unsafe path. Pietro}.
%, as the actions have an uncertain outcome and the state of the robot can change during runtime.
%\todo{Do we mean here mainly that actions have uncertain outcome, or also that the environment itself change at execution time? - Pietro}.
The GP approach, like many other unsupervised learning methods, requires a large amount of evaluations of the policy, making it difficult to run the learning on a physical robot platform, and it can be severely computationally expensive to run in simulation, on an accurate simulator.

In this paper, we also show that for the high-level policy contained in the BT, the detailed physical models of the robot itself, the environment, and the interaction between them may not be necessary, as long as the outcome of different actions is similar enough to the real scenario. We thus propose to learn on a simplified simulator with a low computational cost, and show that the solutions found are still valid for solving the task in a highly detailed simulation.

The paper is organised as follows. Section~\ref{sec:work} introduces Behavior Trees and Genetic Programming and an overview of methods to synthesize BTs. In Section~\ref{sec:method} we present our approach and demonstrate its performance in Section~\ref{sec:experiments}.
%\todo{Can we claim the in none of the above works the environment in uncertain? I would also suggest to emphasize that in robotics dealing with uncertainties is of crucial importance, differently from most videogame applications - Pietro}

\section{Background and Related Work} \label{sec:work}

%\begin{itemize}
%    \item Learning BT: describe various approaches focusing on learning the structure
%    \item Learning BT: describe problem and complications (- computation time)
%    \item state how we improve SOTA
%\end{itemize}
%\todo{I think that this section should start by describing all previous literature, along with pros/cons, and then end with a description of how your work relates, so that you get a single summary of your approach, rather than having it spread out among the related work - Christian - check}

In this section we provide a background on Behavior Trees and Genetic Programming, and a summary of the related work, and show how our proposed methods addresses shortcomings in the state of the art.

%Note that the problem of learning the structure of a BT limits the learning methods that can be used, due to the fact that the search space is unlimited: even if the list of possible nodes is limited, the size of the tree is not. \todo{well technically we could limit the size of the tree.}\todo{I don't think that there is any a priori known upper bound on the size of the BTs that is guaranteed to contain a working solution - Christian}

\subsection{Behavior Trees}
%\todo{Add a short description of the specific properties of BTs that are relevant when using GPs, such as reuse of subtrees, etc - Christian - check}

Behavior Trees (BTs) are a policy or controller representation for AI agents originating in the gaming industry, later applied to robotics, seeing use as an alternative to Finite State Machines (FSM)~\cite{colledanchise_behavior_2018}. BTs have natural support for task hierarchy, actions sequencing and reactivity, and improve on FSM especially in terms of modularity and reusability~\cite{iovino_survey_2020}.

\begin{figure}[htbp]
\centerline{\includegraphics[width=0.5\textwidth]{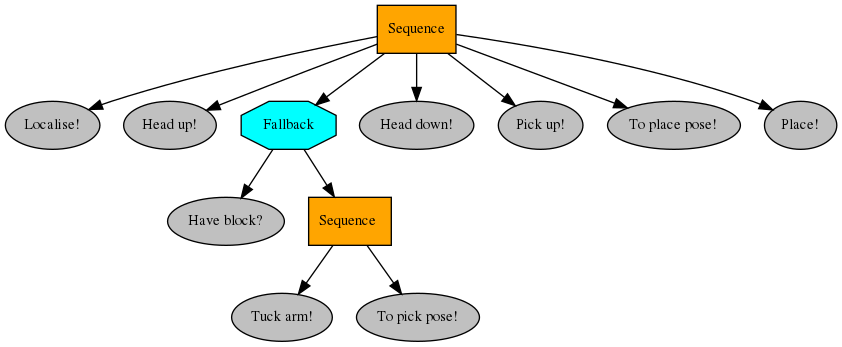}}
\caption{Example of BT with the pool of actions.}
\label{fig:bt_solution}
\end{figure}

In a BT internal nodes are called control flow nodes (polygons in Figure \ref{fig:bt_solution}) and can be of type \textit{Sequence}, \textit{Fallback} or \textit{Parallel}, while the leaves are called execution nodes and can be either \textit{Actions} or \textit{Conditions} (ovals in Figure \ref{fig:bt_solution}). Execution is realised by propagating \textit{tick} signals from the root at a given frequency. Action nodes are executed when ticked and returns one of the statuses \textit{Running}, \textit{Success} or \textit{Failure}. Condition nodes perform status checks or sensing and return \textit{Success} or \textit{Failure}. Sequence nodes execute its children in sequence, returning when all succeed or one fails. The Fallback (or Selector) node, executes its children in sequence, returning when one succeeds or all fail. The return state \textit{Running} is crucial for reactivity~\cite{colledanchise_behavior_2018}, allowing other actions to preempt non-finished actions.

The modularity of a BT is particularly relevant when using evolutionary algorithms such as GP because any leaf or subtree is a building block that can be added to the gene pool and re-used in following generations. In crossover, parents generate offspring by swapping genes sequences and in BTs this can be naturally done with subtrees, without compromising the logic execution of the tree ~\cite{colledanchise_learning_2019} and the safety guarantees~\cite{colledanchise_how_2017}. There is no explicit upper bound on the size of a BT needed to solve a specific task, and this unboundedness of the solution space limits the number of suitable methods to learn new BTs.

\subsection{Genetic Programming}
%\todo{Fill out with some discussion on how the choice of GP mechanisms and parameters affects performance, and/or which versions of GPs are suitable for which kinds of problems - Christian - check}

\begin{figure}[htbp]
\centerline{\includegraphics[width=0.45\textwidth, height=8cm]{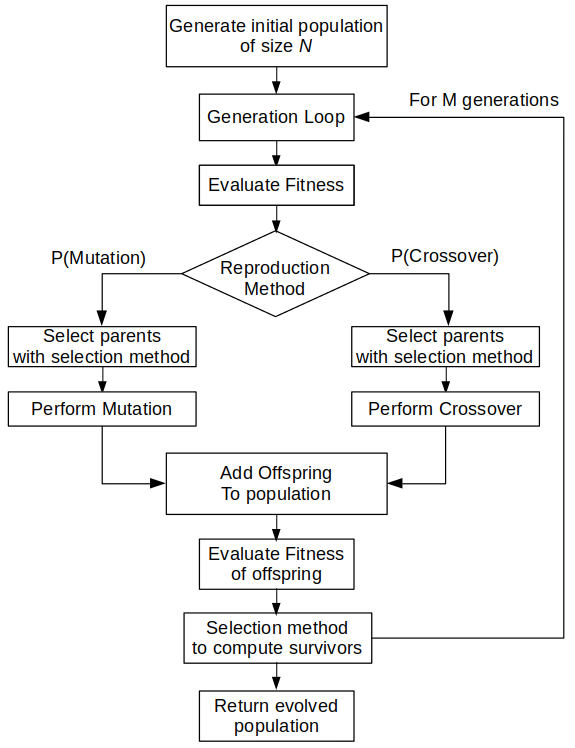}}
\caption{Scheme of GP execution flow}
\label{fig:GP_scheme}
\end{figure}

Genetic Programming (GP) is an optimization algorithm that can evolve programs represented as trees~\cite{koza_gp_1970}. Populations of individuals (computer programs) generate offspring through the functions of \emph{crossover} and \emph{mutation}. A selection mechanism decides which part of the population to keep (Figure \ref{fig:GP_scheme}). Types of selection mechanisms are e.g. \emph{elitism} (keeping the highest ranking), \emph{tournament selection} (individuals are compared pare-wise), \emph{rank selection} (the probability to keep an individual is proportional to its rank in the population). The survival selection is based on a fitness function that assigns a score to each individual based on how it performs in solving the task. There are many variations of GP, in which a grammar is defined to enforce a particular structure and constrain the evolution (Grammatical Evolution), or where only programs parameters are allowed to change and the genotype is represented as fixed-length strings of integers (Genetic Algorithm) \cite{sinclair_evolving_1997}.

\subsection{Related Work}
%\todo{Add some comments on the alternatives to GPs, what other methods exist to automatically generate BTs, and what are their shortcomings - Christian }
%\todo{May want to take a look at this paper: "Evolving Instinctive Behaviour in Resource-Constrained Autonomous Agents Using Grammatical Evolution" Ahmed Hallawa, Simon SchugGiovanni, Gerd Ascheid. EvoApplications 2020: Applications of Evolutionary Computation pp 369-383 - Christian - check}
%It is important to be able to automatically synthesise BTs, both as a tool to help designers to evaluate hand coded solutions, and to help inexperienced users to design a policy to solve a task. 
The two main types of automatic synthesis of BTs are planning and learning~\cite{iovino_survey_2020}. An outcome of a planning algorithm is often a sequence of actions to achieve task goals. Failing to execute an action requires re-planning, so the more uncertain the environment is, the more time and computational resources are devoted to compute new plans to adapt to the situation. A common way to deal with environment changes is to include pre- and post-conditions, exploiting the natural reactivity and modularity of the BT~\cite{colledanchise_towards_2019}. This method exploits back-chaining, starting from the goal conditions and subsequently linking the actions that satisfy them and proceeding backwards, iterating until the starting state is reached. Task knowledge is required to include the conditions. Whenever a post-condition is not met, the BT is expanded to include the actions that achieve it, requiring to stop the execution flow. Updating the BT online during task execution makes it more difficult to analyze and evaluate the policy before it is deployed.

Learning from demonstration and evolutionary approaches~\cite{iovino_survey_2020} have also been proposed as synthesis methods. Learning from demonstration allows inexperienced users to teach a robot how to solve a task. In~\cite{french_learning_2019}, humans teach a house cleaning task to a mobile robot. Primitives are learned from demonstration and incorporated to a Decision Tree, which is translated to a BT. Depending on task complexity, multiple teaching sessions are required to learn the necessary actions. In particular, it is difficult to handle faults that were not encountered during the demonstrations.

Evolutionary approaches have been used to learn BTs in computer game applications~\cite{hutchison_evolving_2010}. The authors raise the problem of execution time, since all BTs had to be evaluated by playing the game DEFCON, and the total execution time was roughly 41 days.
Various methods have been used to learn BTs for the Mario AI Competition~\cite{colledanchise_learning_2019,perez_evolving_2011,nicolau_evolutionary_2017,zhang_behavior_2018,zhang_learning_2018,mcclarron_effect_2016}.  \cite{perez_evolving_2011,nicolau_evolutionary_2017} use Grammatical Evolution (GE). This requires a syntax for all possible solutions in a context-free grammar. This approach requires domain knowledge and an engineering effort to design the grammar, which grows in complexity with the task to solve. Moreover, grammars defining large BTs can become difficult to read and interpret. Grammars can also compromise the logic correctness of the BT structure, requiring ad hoc modifications~\cite{hallawa_evolving_2020}.
In~\cite{zhang_behavior_2018,zhang_learning_2018}, structural and dynamic constraints (i.e. recurrent subtrees are identified during learning and protected from changes) are implemented in the GP, speeding up the learning by preventing the generation of invalid trees. Similar constraints are proven to improve the fitness by not producing undesirable BTs~\cite{mcclarron_effect_2016}.
In~\cite{colledanchise_learning_2019}, a GP is combined with a heuristic that tries all actions until it finds one that improves the reward. If the heuristic fails, then GP is used to combine sequences of actions. The GP operation of mutation is limited to change a node to another of the same type (control or leaf). All available conditions are already included from start and after the learning process an algorithm is run to reshape the solution. In computer game applications, treating uncertainties is not of  importance, since actions have deterministic outcomes.

%In robotics, learning BTs by GP has been proposed for robot swarms. 
In~\cite{neupane_learning_2019} GE is used to evolve a BT to solve a foraging task for a robot swarm. A grammar is used to constrain the evolution of the trees but can push the algorithm to find complex structures. Constructing the grammar requires domain knowledge, and risks overfitting. Authors in~\cite{jones_two_2018} use GP to evolve BTs to control a swarm of robots. Even though they do not publish the resulting BTs, they detail the simulator used for the experiments: the simple structure and interactions for these robots enable fast simulation, including physics and sensing, with an average time of 6 minutes to evolve a population of 32 individuals for 100 generations.

In our approach, we use some structural constraints from~\cite{mcclarron_effect_2016}, 
%\todo{I think it's actually better to cite "Effect of Constraints on Evolving Behavior Trees for Game AI" here?}
to avoid to have the same type of control node on two consecutive levels of the tree, conditions on the rightmost position of any sub-tree, control nodes without children, or having identical condition nodes to be next to each other, as none of these can affect the outcome of an execution of the BT.
%\todo{A couple of other rules implemented is: No control nodes without children, no two identical condition nodes next to each other} 
%Post-conditions would check if actions before them have done the job and restart the tree otherwise, but since the BT is reactive, this mechanism is naturally satisfied.  %\todo{Explain why - short sentence - Christian}.
Unlike~\cite{colledanchise_learning_2019}, we do not constrain the mutation only to nodes of the same type, thus increasing the diversity. We do not explicitly specify which conditions to use, but leave this to the GP algorithm to find. Conditions always required to perform a specific action (e.g. only attempt to pick objects not already held in hand), are included within the behavior at an atomic level. We do not perform explicit reshaping of the tree, but the size of the tree (in terms of number of nodes) is a factor included in the fitness function.

To conclude, our contribution with respect to the state of the art is that our GP algorithm can reach convergence in a non-deterministic robotic scenario without the need for task specific heuristics. This is achieved by synthesizing BTs which are naturally reactive to failures and including the success probability in the fitness function, driving towards solutions that avoid actions that are more likely to fail. %\todo{Please Christian/Jonathan, write this better and clearer. - Matteo}
%\todo{I feel we do not not fully emphasize and sell, at a high level, what is our unique contribution compared to all the above works. For example,  we are able to handle uncertainties in a more efficient way  ? - Pietro }

\section{Proposed Method} \label{sec:method}

Synthesising fault tolerant policies for robotics with unsupervised learning requires the agent to attempt to solve the task in the environment in which it operates, which may require many learning episodes.
This may require significant time, as it is not possible to speed up the execution of real tasks. Safety plays another important role, because in order to learn not to fail, the robot may have to experience failure, with the risk of causing unwanted damage or hurting humans. Finally, resetting the task environment for subsequent runs may often require expensive human intervention.
%\todo{I would say that resetting time is another MAJOR factor here because it's often done manually}
Thus, it is better to first learn the policy in a simulator and then try the learned policy in the real world. In detailed simulators, both kinematics and dynamics can be modelled, together with collision boxes for contacts computation. This allows to realize complex robotics tasks, including sensing, navigation, manipulation and so on. The more complex the models and the task to solve, the higher the computational cost will be. In this perspective, simulators can solve the safety problem, but time is still an issue. To overcome this limitation, we propose to learn the structure of a BT using a very simplified simulator, that does not take details of sensing, kinematics, and dynamics, into account. With a careful design of the simulator, the high-level decision and the execution flow would be the same. We demonstrate this with a state machine standing in as a simulator, but assume that any simulator that keeps the high-level structure of actions and outcomes of the original task would work.

%We will here describe our proposed GP learning algorithm, and and example task to be solved with a BT.

%orm of GP, of a policy, represented as BT, to solve a mobile manipulation task in which the actions have an uncertain outcome. The same learning problem has also been attempted on the more detailed physical simulator Gazebo, a state of the art for ROS applications. In this section we will describe our implementation of the GP operations on BTs, then the designing principles of the state machine will follow.
%\todo{Start with describing the original problem, learning on a detailed simulator (or real world), and then move on to describe the problems (time consumption). Explain why the state machine is a good idea. What requirements would we expect to place on it in order to work as a stand-in for the real problem? What would we expect from the SM solution in terms of being applicable to the real world (simulator) problem? e.g. high-level decisions should be the same, but details of mechanical interaction will not be covered, etc...   -Christian}

%\begin{itemize}
%    \item describe method and differences with  \cite{colledanchise_learning_2019} and \cite{zhang_learning_2018}
%    \item describe how we dealt with probabilities and differences with \cite{colledanchise_performance_2014}
%    \item describe State Machine as Markov Model
%    \item reward design shaping
%\end{itemize}

\subsection{GP algorithm and operations on BTs}
\begin{figure}[htbp]
\centerline{\includegraphics[width=0.4\textwidth]{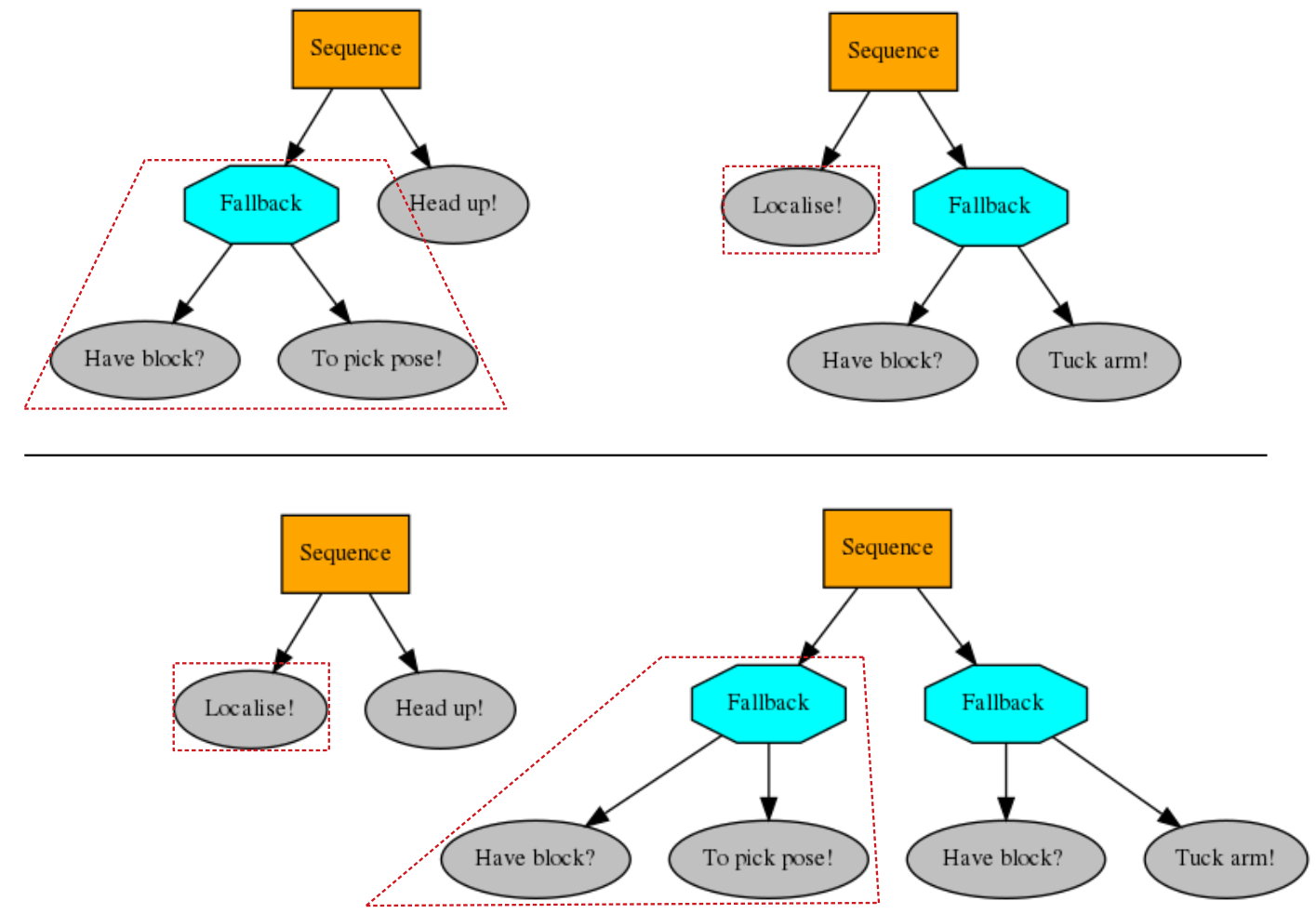}}
\caption{Example of a Crossover operation in BTs. Top: parent generation, bottom: offspring generation.}
\label{fig:crossover}
\end{figure}

The gene pool provided to the genetic algorithm, i.e. the \textit{primitive set} of the GP system, is composed by a set of \textit{terminals} (the BT leaves: actions and conditions) and a set of \textit{functions} (the BT control nodes: sequence (s) and fallback (f)). The BT is represented as a string (e.g. [`s(', `action1', `action2', `)']), where the parenthesis are used as subtree identifier. The GP functions operate on the string, which is mapped into a BT for the evaluation. The initial population is composed of $N$ randomly created BTs of length 4 and then it evolves using the following functions:

\paragraph{Crossover} this function takes two individuals of the parent generation and performs a sub-tree swapping, returning two offspring individuals, as in~\cite{colledanchise_learning_2019} (Figure \ref{fig:crossover}).

\paragraph{Mutation} this function changes an individual in three different ways with set probabilities, keeping the structure of the BT consistent. For node addition and mutation, there is a 50\% chance that a control node will be chosen, since they typically constitute around 50\% of the final behavior tree.
\begin{enumerate}
    \item Node mutation: a node in the individual is mutated to any node in the gene pool.
    \item Node addition: a node from the gene pool is added to the individual at any level.
    \item Node deletion: a node is removed from the individual.
\end{enumerate}

\begin{table}[htbp]
\caption{GP parameters used in the simulations}
\begin{center}
\begin{tabular}{|c|c|}
\hline
\multicolumn{2}{|c|}{\textbf{GP Parameters}} \\
\hline
individuals in population ($N$) & 30  \\
individual start length & 4 \\
generations & 8000 \\
crossover \% & 40 \\
mutation \% & 60 \\
elitism \% & 10 \\
mutation: node mutation \% & 30 \\
mutation: node addition \% & 40 \\
mutation: node deletion \% & 30 \\
selection method & tournament \\
\hline
\end{tabular}
\label{table:sim_pars}
\end{center}
\end{table}

For the \emph{Crossover} and \emph{Mutation}, the parents are chosen by running a \textit{Tournament Selection}, giving the number of individuals shown in Table~\ref{table:sim_pars}, as a percentage of the population. Tournament Selection runs several single duels in which the most fit individual of the pair survives. For a given generation, the individual with the best score will always survive, while the individual with the worst score will not. For the rest, it depends on the fitness of the individual they are paired with. Tournament Selection generally performs better than \textit{Elitism}, in terms of convergence~\cite{shukla_comparative_2015}, because it allows keeping genetic content used in poorly performing individuals, resulting in a better exploration of the search space and in a lower chance of getting stuck in local minima.
%\todo{Maybe a sentence about this being because it's better at exploring and doesn't get stuck in local minima as easily?}
%\todo{Do you have any results showing the performance difference between tournament and elitism? - Christian}

Crossover and mutation will generate two different offspring for each parent, to keep the population size low without compromising the learning speed. The crossover function is repeated twice for each pair of parents, while making sure that there are no copies in the offspring. For mutation, a single parent will generate two different offspring according to the mutation function.
A percentage of the following generation (`elitism' in Table \ref{table:sim_pars}), is populated by the most fit individuals. The remaining slots are populated by the output from another Tournament Selection run with all the parents and the offspring ($N+2N$ individuals).

\subsection{Simulator design}
The simple simulator we use here is based on a state machine, where the states and the transitions are designed to be as close as possible to the realistic physics simulator, also providing the same kind of feedback: e.g. estimated robot pose from the localization filter, robot configurations, object pose, etc. The transitions depend on the state of the robot and on the outcome of the action being ticked. The transitions can be designed to be deterministic or probabilistic, modelling the simulator as a Markov Chain. To the best of our knowledge, there is no work leveraging this kind of simulator to learn a BT. In the literature, learning of the structure is always studied in a deterministic fashion (e.g. computer games). In GP every individual needs to be scored by the fitness function. Every BT designed by the GP is simulated in the physical simulator, making the problem intractable if the number of individuals in a population or the number of generations is high (in the order of magnitude of hundreds). In our state machine simulator each evaluation is several orders of magnitude faster.
Note that the state machine is meant to substitute the physical simulator. Thus, it is seen as a black box by the learning algorithm, taking as input an individual and returning the fitness score. The knowledge required to design the state machine is not incorporated in the learning algorithm. Since the outcome of the actions is uncertain, post-conditions cannot be defined, thus invalidating planning approaches as in \cite{colledanchise_towards_2019}. 
%\todo{highlight that the SM simulator also evaluates every individual, but must faster - Christian}

\section{Experiments and Results} \label{sec:experiments}

%\begin{itemize}
%    \item Contribution 1:
%        \begin{itemize}
%            \item solution with different levels of uncertainty
%            \item solution with different level of knowledge on the task (if it converges)
%            \item solution with more complex task (more objects?)
%            \item show that solutions work in the Gazebo simulator
%        \end{itemize}
%    \item Contribution 2:
%        \begin{itemize}
%            \item detail Gazebo solution in terms of characteristics of the HW/SW and exec time
%            \item detail SM solution and test it in Gazebo (video + images)
%        \end{itemize}
%\end{itemize}
%\todo{Introduce what and whay you are doing, e.g.: "In order to evaluate the performance of our approach, we test it on a mobile manipulation task" - Christian}

\begin{figure}[htbp]
\centerline{\includegraphics[width=0.4\textwidth]{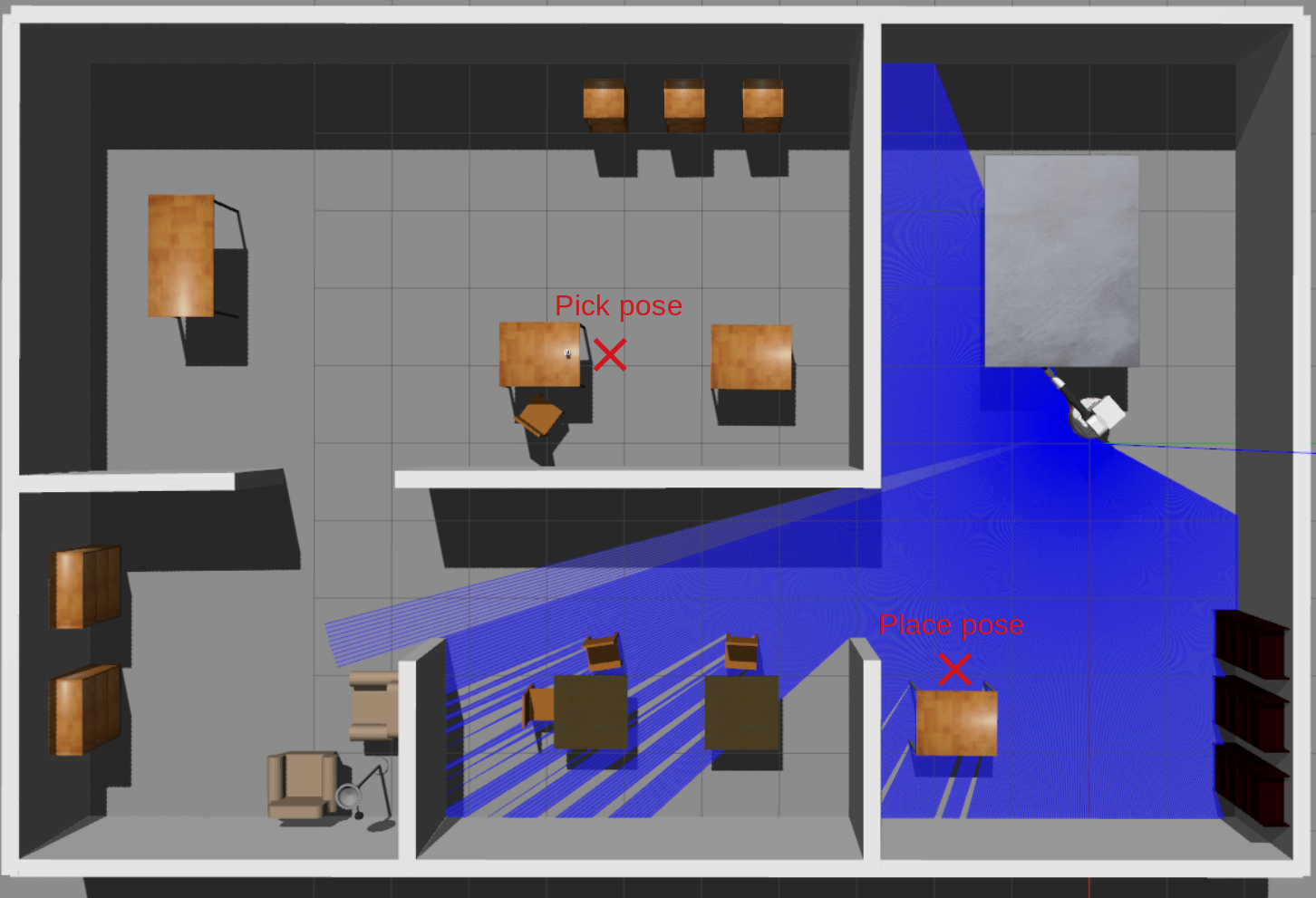}}
\caption{Simulation environment in Gazebo.}
\label{fig:environment}
\end{figure}

To evaluate the performance of our approach, we test it on a mobile manipulation task. The task to solve is for a mobile manipulator to pick an object and place it in another position, thus requiring both navigation and manipulation.
To navigate the environment, the robot must be localised and its arm needs to be in a tucked configuration, to not collide with the obstacles. The robot has a camera for object recognition located in the head, which has two configurations: \textit{Up} (to navigate) and \textit{Down} (to identify and manipulate objects). To complete the task the robot has to place the object in a goal position. The object is assumed to be a cube, and the pick/place positions correspond to two different tables in the environment (Figure \ref{fig:environment}).
The set of action behaviors available to the robot are are \textit{Localise}, \textit{HeadUp}, \textit{HeadDown}, \textit{Tuck}, \textit{Pick}, \textit{Place} and \textit{MoveTo}, which can then take as input parameter a goal pose (e.g. the pick and the place pose). A condition \textit{Have Block?} checks if the robot is currently holding the cube. This is the set of the necessary behaviors to complete the task (Figure \ref{fig:bt_solution}). %Other behaviors will be added to the problem to assess the algorithm robustness to noise, and to simulate a realistic case where more options are available to the robot than just those strictly necessary to solve the task.

\subsection{Fitness function design}

For the BTs to evolve, a trade off must be made between generalization and domain knowledge in the fitness function. If it is well adapted to the specific task, it will feature a clear gradient that makes the learning faster, however it may be too task specific and difficult to generalize to other tasks. However, using a very generic fitness function, such as a binary ``fail/success" might fail due to the lack of gradient, making the problem closer to a random search. Taking this into account, we design the fitness function to include all the elements that are common to mobile manipulation tasks. %and to constrain the structure of the tree, favouring smaller trees to improve readability.
%\todo{explain why this is a good idea - Christian}.
The fitness function is an objective to maximize:
\begin{equation}
    \mathcal{J} = \arg \max (- \mathcal{C}).
\label{eq:fitness}
\end{equation}
The function $\mathcal{C}$ is the cost:
\begin{equation}
    \mathcal{C} =  f(s, b) + \gamma T + \delta P ,
\label{eq:cost}
\end{equation}
where 
\begin{equation}
    f(s, b) = \alpha \mathbf{s}^2 + \beta b,\ \mbox{and}\  s = || \mathbf{s_d} - \mathbf{s} ||
    \label{eq:f}
\end{equation}
%\todo{ And what is f()?}
takes into account the distance of the robot and cube poses (states $s$) from the desired goal pose $s_d$. In particular, we award the robot being close to the cube, to provide a ``hint" that this may help solve the task. It also includes the localization error, measured as the distance between the real pose of the robot in the environment and the estimated one. The vector weight $\alpha$ for the distance factor has components $\alpha_1$ for the distance cube-goal, $\alpha_2$ for the distance robot-cube and $\alpha_3$ for the localization error. The term $b$ takes into account the length (the number of nodes) of the tree, $T$ is the execution time and $P$ is the estimated failure probability, obtained by summing the failure probability of the executed actions. \cite{colledanchise_performance_2014}~provides a tool to compute the prior estimation of the success/failure probability of the whole tree, but here we take into account the estimated failure probability despite the outcomes of the actions. In this way, the cost increases when actions with high probability of failure are executed. Finally, we give a reward when the robot achieves the picking of the cube and when it finally places it. This in particular prevents the robot from stopping with the cube in front of the placing pose but pushes it to complete the task. The fitness function parameters are in Table~\ref{table:fitness_pars}.
%\todo{Adding a sentence to make this important concept more clear. Maybe something like: This way, the cost associated to a behaviour tree increases when actions with high probability of failures are executed. - Pietro}

\begin{table}[htbp]
\caption{Parameter for the functions in (\ref{eq:cost}) and (\ref{eq:f}).
%\todo{connect to equation (2)}
}

\begin{center}
\begin{tabular}{|c|c|}
\hline
\multicolumn{2}{|c|}{\textbf{Weights}} \\
\hline
pick reward & 50  \\
place reward & 100 \\
$\alpha_1$ (distance cube-goal) & 10 \\
$\alpha_2$ (distance robot-cube) & 2 \\
$\alpha_3$ (localization error) & 1 \\
$\beta$ (BT length) & 0.5 \\
$\gamma$ (execution time) & 0.1 \\
$\delta$ (failure probability) & 0.0 \\
\hline
\end{tabular}
\label{table:fitness_pars}
\end{center}
\end{table}

Completing the entire task, or the sub-task of picking the cube are important, and given high weights, while less important goals that are intended to provide gradients for partial solutions are given significantly lower weights. $\delta$ is not used in the first experiment.

%The choice of the parameters is driven by the willing to push the robot to complete the task (`distance cube-goal'), giving it small hints on how to do so (`distance robot-cube' and `localization error'), but high rewards for task milestones (pick and place rewards). The weights for the BT length and the execution time are chosen to give a secondary objective to the GP algorithm: among the individuals that solve the task, favour those with a small structure and that take less time to execute (and minimize the failure probability if $\delta \neq 0$).

\subsection{Experiments}
%\todo{For each experiment, add comments about how the found solution looked, that it worked in Gazebo, etc.  - Christian}
The task has been designed in the physical simulator Gazebo, with the robot controlled through ROS (Figure \ref{fig:environment}). We use this to evaluate the performance of BTs learned using the simple simulator.

We also tried running the GP directly in this simulator, but it takes more than a month of runtime on a powerful gaming computer % with a 16 GB RAM, a 12-core Intel i7-850H at 2.20 GHz and a NVIDIA GeForce RTX 2070,
 to converge.
% for the graphics, the algorithm still did not converge. In addition, we speeded up the process both by parallelizing the simulations, using four Docker containers running on 4 cores each, and by reducing the number of simulations to the unseen BTs, storing in an hash-table each BT with its fitness score. We also considered to remove the dynamics from the simulator, limiting the computation to the kinematics and the collision checking, but the amount of the engineering required led us to drop this option.
%\todo{ As a reviewer I would ask here: What if we use a simulator with only kinematics/collision-checking functionalities and no dynamics? Would this require less work than with a state machine and would it be also faster than dynamical simulation? - Pietro}
%\todo{So should we really blabber this much about something that didn't work? Is this really interesting?} \todo{In this case Christian will prune. - Matteo}
The same learning problem, set up to run in the simple state machine simulator, allows both to speed up the learning by several order of magnitudes, down to a few minutes (depending on the GP parameters), and to choose the failure probabilities and outcomes for all actions.% What we mean is that it is up to the designer how to model uncertainties and we get rid of sudden system failure that are beyond the simulation and likely to happen in detailed simulators (e.g. ROS packages dropped, navigation/manipulation planners failure).

%\todo{Shouldn't be to hard to just simulate failures with a given probability though?}

%Using the state machine as simulator, 
%We demonstrate 3 things in our experiments. 

\begin{figure}[htbp]
\centerline{\includegraphics[width=0.4\textwidth]{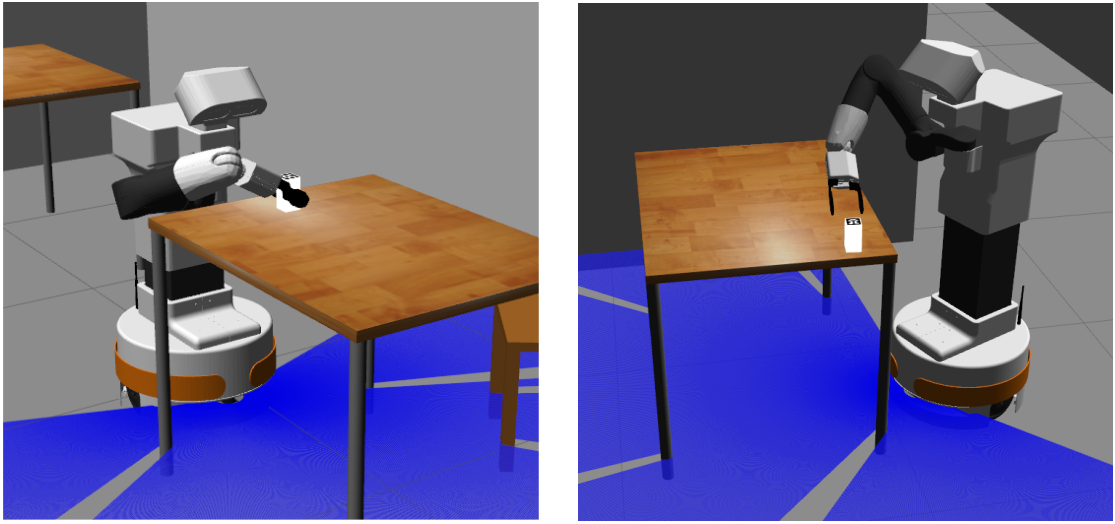}}
\caption{Pick and Place in Gazebo.}
\label{fig:pick_place}
\end{figure}

%We also demonstrate that the learning method is robust to noise, by adding behaviors in the gene pool which are not required in order to complete the task. Then, we will manipulate the weight $\delta$ in Eq. \ref{eq:cost}, to show how the solution will vary with respect to the risk we are willing to take.
All the experiments are carried out with the GP parameters in Table \ref{table:sim_pars}. The BT can fail up to 5 times, below this limit, the root is ticked again upon failure, so that the robot can perform again an action that just failed. This approach is logically equivalent to have copies of the same action controlled by a Fallback node, but results in a simpler tree structure.
%\todo{explain why we do like this}
Every learning curve is an average over 10 runs. The learned BT solutions in all the following experiments were verified to also solve the task in the more detailed simulator in Gazebo. (See Figure~\ref{fig:pick_place}).

In experiment~1, we will show how the failure probability affects the learning of a robust BT. For this we first run a simulation on a deterministic state machine, where the actions do not have a failure probability. Then we run four more simulations with increasing levels of uncertainties, as showed in Table \ref{table:comparative}. 

\begin{table}[htbp]
\caption{State Machine values for the failure probabilities}
\begin{center}
\begin{tabular}{|c|c|c|c|c|}
\hline
\multicolumn{5}{|c|}{\textbf{Failure Probability}} \\

\hline
Probability Type & Stoch. 1 & Stoch. 2 & Stoch. 3 & Stoch. 4 \\
\hline
localization failure & 0 & 0 & 0.2 & 0.3  \\
pick failure & 0 & 0 & 0.2 & 0.4 \\
place failure & 0 & 0 & 0.1 & 0.2 \\
losing cube & 0 & 0.05 & 0.05 & 0.1 \\
losing localization & 0.1 & 0.1 & 0.1 & 0.2 \\
\hline
\end{tabular}
\label{table:comparative}
\end{center}
\end{table}

\begin{figure}[htbp]
\centerline{\includegraphics[width=0.5\textwidth]{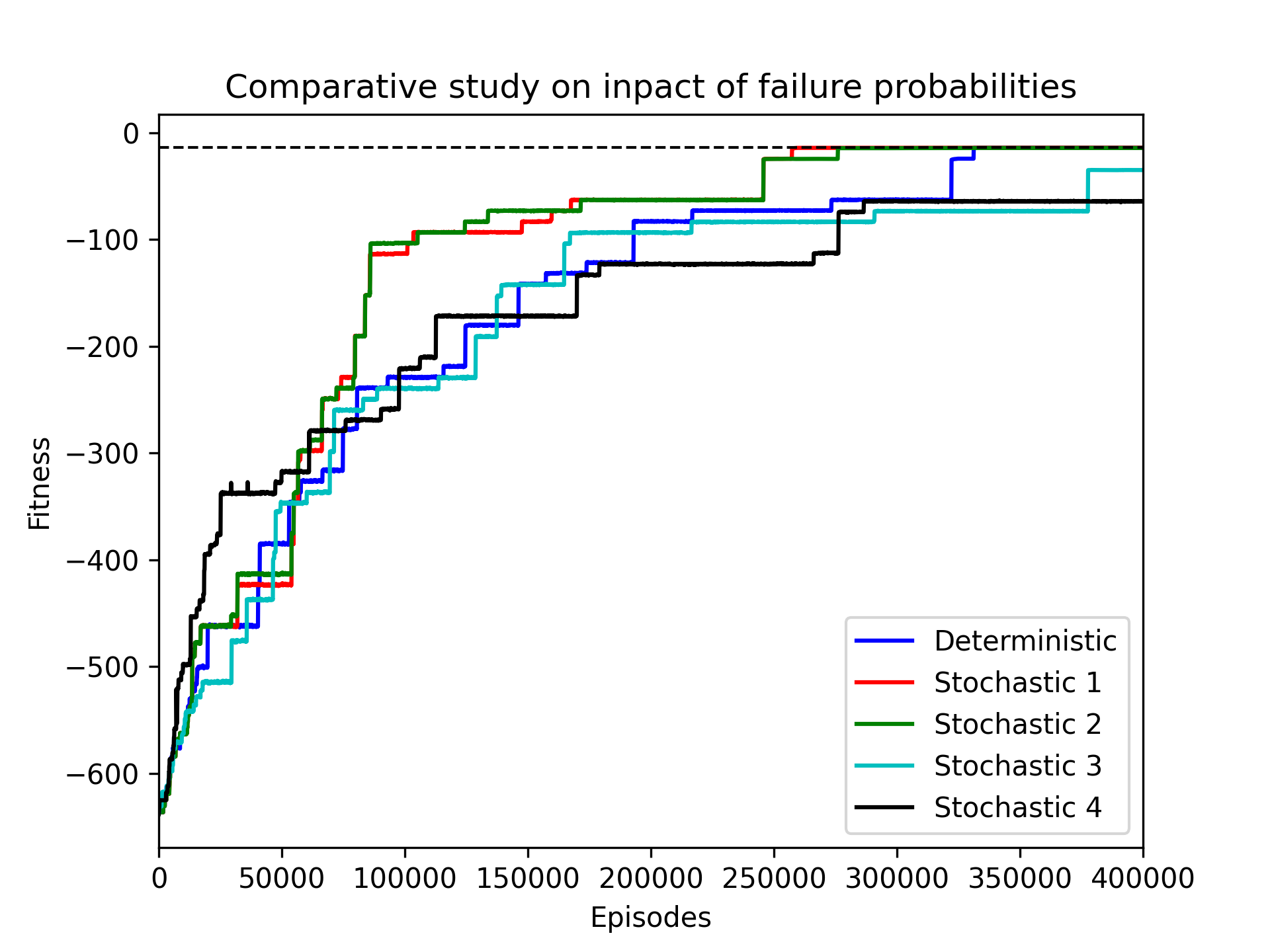}}
\caption{Effect of failure probabilities on learning, averaged on 10 runs.}
\label{fig:failurecomp}
\end{figure}

The `localization failure' is the probability of failure in the convergence of the particle filter that estimates the robot pose, whereas the `losing localization' is the probability for the robot to lose the localization during the navigation. `Losing cube' means dropping the cube when the robot moves. When the cube is dropped, it will respawn to the pick pose, thus requiring the robot to pick it again. The third column will be used as baseline for comparison in the remainder of this section. The results are shown in Figure~\ref{fig:failurecomp}.

The algorithm is able to handle uncertainties and grant the same convergence as in the deterministic case. When the level of uncertainty increases, however, the found solution is generally more complicated and features multiple copies of the same action, to increase the overall success probability.

Note that the number of episodes needed for convergence, approx. 400 000, would be expensive to run on a real robot or on a realistic simulator. An example of the output is Figure~\ref{fig:bt_solution}.
% deterministic_2 and stoch4_3

\begin{figure}[htbp]
\centerline{\includegraphics[width=0.5\textwidth]{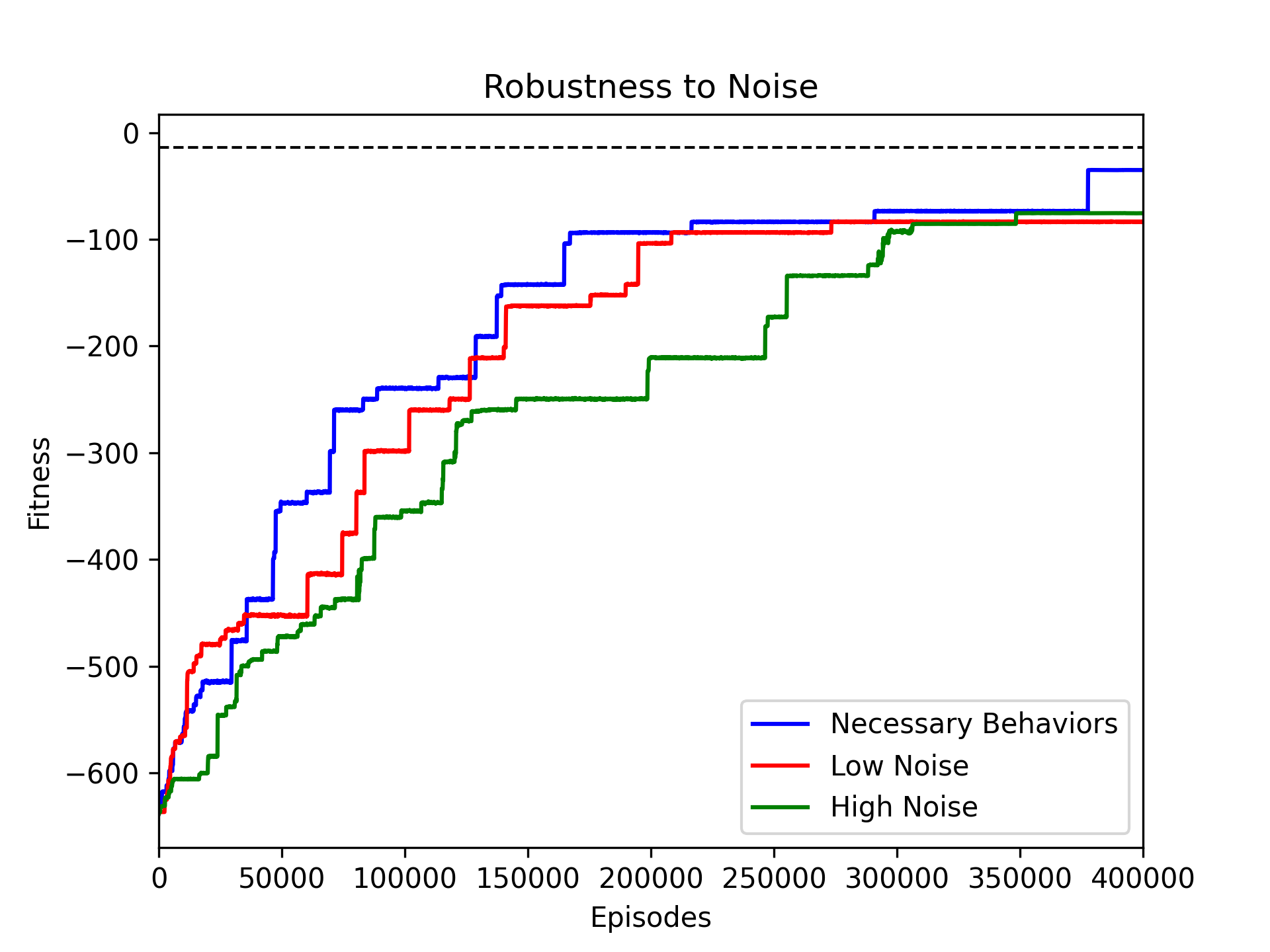}}
\caption{Effects of noise on the learning.}
\label{fig:noise}
\end{figure}

In experiment~2, we compare a setup in which the pool contains just the 9 behaviors necessary to the completion of the task to two noisy setups. In the first comparison we add 3 useless behaviors (\emph{low noise} in Figure \ref{fig:noise}), while in the second we add an additional 27 (\emph{high noise}). This is to emulate the case where the robot has access to a large number of actions, but only a small subset are needed for a particular task. %\todo{There must be a way to describe this more simply, even I don't understand exactly what you mean here /Jonathan}

The results shown in Figure \ref{fig:noise} show that the GP is robust to added unnecessary actions. The added actions could have led the robot to local minima, being mostly~\textit{MoveTo} type behaviors that drive the robot around the environment and could get either close to the cube or close to the goal position. The GP excludes the meaningless behaviors in early stages.

\begin{figure}[htbp]
\centerline{\includegraphics[width=0.5\textwidth]{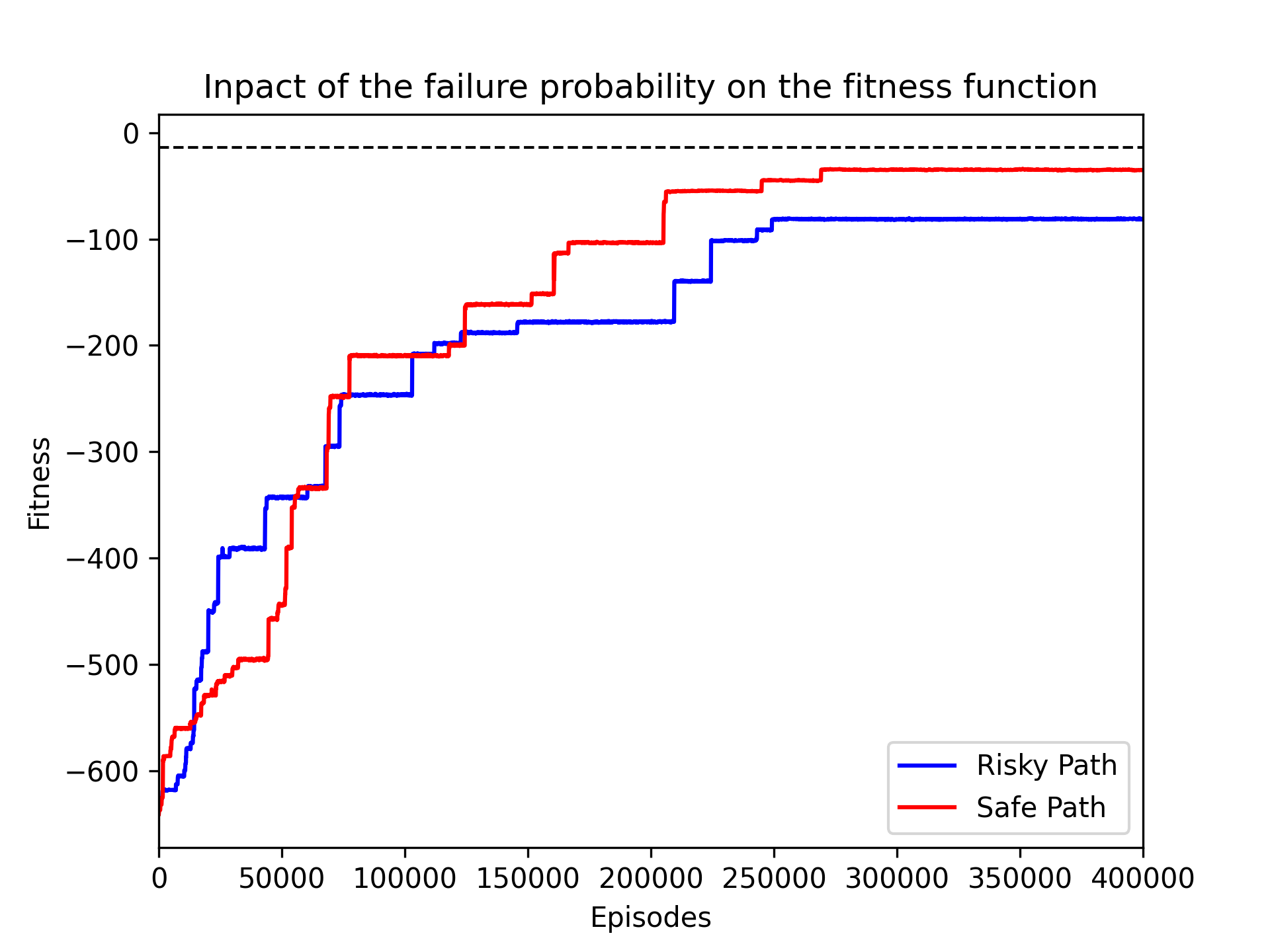}}
\caption{Effects of the $\delta$ parameter on the Fitness function.}
\label{fig:runs_safe}
\end{figure}

In experiment~3 we explore how the parameter $\delta$ affects the learned solution. We assume that the robot could take two paths to both the pick and the place positions: a shorter one, but cluttered by human presence, thus with a higher risk of losing the cube or the localization (set to 0.2 and 0.4 respectively), and a longer one without these probabilities but a longer execution time. This is achieved by setting the \textit{MoveTo} behaviors to take the shorter but more risky path and by adding \textit{MoveTo*} actions taking the other. The results are reported in Figure \ref{fig:runs_safe}, where the $\delta$ parameter is set first to 0 and then to 150, thus penalizing the robot for taking risks. %Pushing the robot to take safer version of the same behaviors improves the convergence.

\begin{figure}[htbp]
\centerline{\includegraphics[width=0.5\textwidth]{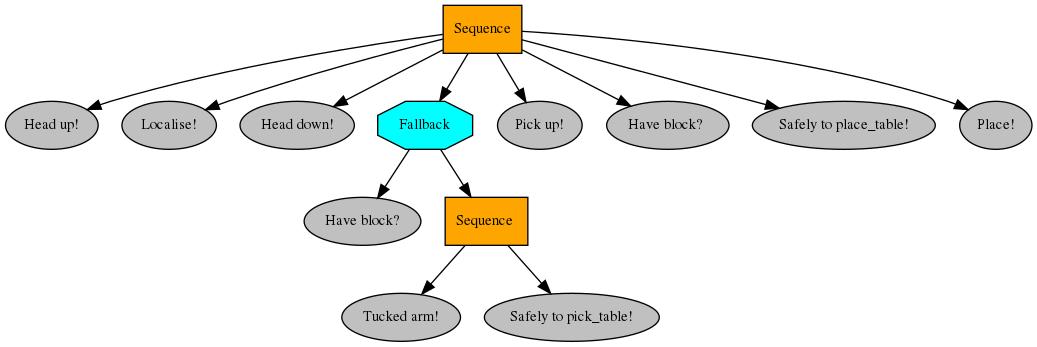}}
\caption{BT taking safer paths to pick and place.}
\label{fig:BTsafe}
\end{figure}

Figure \ref{fig:BTsafe} shows that the same BT structure that solves the task in normal conditions (see Figure \ref{fig:bt_solution}), now features the modified \textit{MoveTo*} behaviors taking the safer path. 

% \todo{are these results presented somewhere?} \todo{we will include a video - Matteo} %This requires the same pool of behaviors to be implemented. If also the same algorithm to map strings to BT is available, the same solution string can be executed in both simulators with similar outcomes.
%For cases where the solution would not work directly in the advanced simulator or on the real robot, the BTs output by the proposed method could be used to bootstrap the learning, rather than restarting with a set of random BTs. %To this extent, the final population that evolved by GP on the state machine can be used to seed the GP on the physical simulator and run for few more runs, in order to refine the solution.

\section{Conclusions and Future Work} \label{sec:conclusion}

%\begin{itemize}
%    \item stress importance of State Machine approach, in particular when it allows to control stochastic nature of the task
%    \item recap
%    \item future work: comparison with student solutions and bootstrapping learning on Gazebo with good candidates from State Machine
%\end{itemize}

In this paper we showed that GP can be effectively used to learn the structure of a BT to solve a task in an unpredictable environment. Moreover, we proposed to use a state machine model of the simulator, instead of a physical one, to make the problem computationally tractable, without compromising the traits of the solution. The BTs learned in the simple simulator are demonstrated to efficiently solve the same task in a physical simulator. %We also emphasized the benefit of such approach: many degrees of control on the action outcomes, dynamic and kinematic models not required, contacts not modelled. Once faster physical simulators will be available, the state machine could be potentially abandoned. 

We showed that the learned solution is tolerant to faults on the manipulation, localization and navigation actions, making our method appealing for real robotic applications. As future work we propose to transfer the solution on a real robot setup and to compare it to hand coded BTs. %\todo{This hand coded thing just pops up out of nowhere here.. people that don't have the history might need more info} \todo{I reformulated, but I don't think it's relevant to provide that background. - Matteo}
A natural continuation of this work is to study if learning on a simple simulator can bootstrap the learning in the physical simulator, or real platform for the cases where a solution from the simple simulator does not work directly.% On the same line, we will add the possibility of populating the gene pool with useful subtrees at run-time, as in \cite{zhang_learning_2018}. 
%To conclude, we will investigate on other methods to bootstrap the learning, with Learning from Demonstration as potential candidate to build the first generations for GP, as an alternative to the random approach used here.

\section*{Acknowledgment}

The authors would like to thank Ignacio Torroba Balmori and Christopher Iliffe Sprague for their contribution on setting up and designing the simulation environment in ROS-Gazebo. 

\bibliographystyle{ieeetr}
\bibliography{main}

\end{document}